\documentclass{article}
\usepackage{spconf,amsmath,graphicx}
\usepackage{xcolor}
\usepackage{url}
\usepackage{hyperref}

\usepackage{float}
\restylefloat{table}
\usepackage{multirow}
\title{Large Language Models as User-Agents for Evaluating Task-Oriented-Dialogue Systems}
\name{Taaha Kazi, Ruiliang Lyu, Sizhe Zhou, Dilek Hakkani-Tür, Gokhan Tur 
\thanks{}}
\address{University of Illinois at Urbana Champaign \\
\texttt{\{tnkazi2, ruiliang, sizhez, dilek, gokhan\}@illinois.edu}.}
\begin{document}
%
\maketitle
\begin{abstract}
Traditionally, offline datasets have been used to evaluate task-oriented dialogue (TOD) models. These datasets lack context awareness, making them suboptimal benchmarks for conversational systems. In contrast, user-agents, which are context-aware, can simulate the variability and unpredictability of human conversations, making them better alternatives as evaluators. Prior research has utilized large language models (LLMs) to develop user-agents. Our work builds upon this by using LLMs to create user-agents for the evaluation of TOD systems. This involves prompting an LLM, using in-context examples as guidance, and tracking the user-goal state. Our evaluation of diversity and task completion metrics for the user-agents shows improved performance with the use of better prompts. Additionally, we propose methodologies for the automatic evaluation of TOD models within this dynamic framework. We make our code publicly available \footnote{\url{https://github.com/TaahaKazi/user-agent}}

\end{abstract}

\section{Introduction}
Task-oriented dialogue (TOD) systems are designed to assist users in completing specific tasks or goals through turns of natural language interactions ~\cite{wen-etal-2017-network, liu-etal-2018-dialogue, TODS-review}. 
These systems are typically built to guide users through a series of steps to accomplish a particular objective, such as booking a flight, ordering food, or scheduling an appointment.

Along with the emergence of large language models (LLMs)~\cite{radford2018improving, radford2019language, brown2020language, touvron2023llama1, touvron2023llama}, TOD systems have been fundamentally boosted by the complex reasoning/understanding ability and the domain adaptation power of LLMs. 
The functionality as well as the design concepts of TOD systems have also been tightly integrated into the training and grounding of LLMs.
From the real-world deployment side, at least tens of LLMs have been developed and integrated into hundreds of downstream tasks, for example, the travel planning LLM agents~\cite{Xie2024TravelPlanner}.

However, with such a large number of downstream application scenarios, the capacities of traditional benchmark-based TOD system evaluations are falling far behind. 
This essentially indicates that a gap exists between the TOD system evaluations and the TOD system deployments. 
There are dataset-based evaluation methods where the inputs such as the dialogue schema or the dialogue history are fixed and the evaluation based on the expected output is rigid, which means the evaluation systems cannot evaluate multiple turns of utterances from TOD systems coherently~\cite{cheng-etal-2022-multiwoz}. 
This, however, generally leads to a policy mismatch between the evaluation systems and the TOD systems. 
There are also traditional interactive evaluation methods that generally result in evaluation simulators in pursuit of maximal information exchange~\cite{davidson2023user}. 
Existing LLM-based evaluation methods, however, lack of emphasis on modeling more complex user behaviors~\cite{davidson2023user}.

To tackle the above issues, in this work, we set up our goal as building an automated evaluation system (a user simulator and an evaluator agent) for TOD systems. 
Specifically, given any TOD system, our user simulator agent will simulate conversations with the system given a set of initial user goals and the evaluator agent will output several metric scores gauging the comprehensive performance of TOD systems being evaluated. Figure~\ref{fig:framework} shows our proposed framework that involves the interactions of the user simulator with the TOD system in evaluation and the evaluation module.
The user simulator can state the user goal and associated information, discuss options provided as solutions by the TOD system, and negotiate to find options that match user needs. 
We follow several principles for our simulator design: 
\begin{itemize}
\itemsep -0.5ex
    \item The user simulator should be plug and play, which means it does not need  scenario-specific annotations to fine-tune the simulator and it should be adapteble to different TOD systems or different evaluation domains. 
    \item The user simulator should be rich in linguistic diversity.
    \item The user simulator should be able to model complex user behaviors.
\end{itemize}



In this work, LLMs are used by simulators to evaluate TOD systems. Given an initial user goal, the simulator prompts the LLM to generate responses that interact with TOD systems to achieve the given goal. We propose three prompting strategies: the Vanilla Prompt with straightforward instructions and in-context examples, the Thought Prompt incorporating reasoning steps via Chain-of-Thought, and the User State Tracking Prompt, which updates a user-state tracking dictionary to avoid premature conversation ending or looping.

We use multiple metrics to assess the TOD system's performance, including task completion, naturalness, coherence, and dialogue-level diversity; with task completion evaluated using GPT-4. This approach addresses the limitations of traditional evaluation methods in a dynamic setting. In summary, our work introduces an automated framework using LLMs as user simulators to evaluate TOD systems. This framework enhances the evaluation process, making it more reflective of real-world interactions.

\section{Related Work}

\subsection{User Simulation for TOD System Evaluation}

User simulators are designed to simulate users' behaviors in dialogues, which helps interactive evaluation of TOD systems. The development of user simulators for evaluating TOD systems can be broadly divided into three stages. Initially, user simulators are built based on statistical models~\cite{ai-weng-2008-user, DBLP:journals/air/Lopez-CozarCM06}, mimicking user behaviors that are conditioned on user goals and actions in the context of an interaction. At that time, using user simulators to test TOD systems was rarely explored due to the scarcity of natural and high quality user simulator responses.
More recently, researchers leveraged the power of pre-trained language models (PLMs)~\cite{Devlin2019BERTPO, Radford2019LanguageMA} and developed end-to-end user simulators for TOD system evaluation. \cite{10.1145/3596510, cheng-etal-2022-multiwoz} both build upon the T5~\cite{10.5555/3455716.3455856} language model by training user simulators on existing datasets, such as MultiWOZ~\cite{budzianowski-etal-2018-multiwoz}. Similarly, ~\cite{lin-etal-2022-gentus} uses BART~\cite{lewis-etal-2020-bart} as the basis of the user simulator, which is further refined using reinforcement learning. Though demonstrating impressive goal-success rates (GSR), these user simulators rely on fine-tuning PLMs on annotated, domain-specific dialogue data, which limits their applicability in the testing of new TOD systems.

As LLMx prevail in massive downstream tasks and demonstrate high quality response generation ability, they are employed for user simulator construction as well. Specifically, \cite{DBLP:journals/corr/abs-2110-06905} leverage LLMs as user simulators provided with a list of user goal APIs. \cite{davidson2023user} harnesses the power of in-context learning~\cite{kojima2022large} to generate more diverse and human-like utterances. While achieving great evaluation effects, these systems still suffer from problems such as complex grounding goals and the need to pre-define user goals. Our work attempts to solve these problems by creating a flexible user simulator that does not need fine-tuning for specific tasks and can adapt to different dialogue systems easily. 
We use prompt strategies like Chain-of-Thought and User State Tracking prompts, which ensure logical and complete conversations by guiding the LLM and tracking user goals to avoid early endings or loops. These prompts also help in grounding the conversation.


\subsection{LLMs for Task Completion}

Traditional LLM benchmarks like MMLU~\cite{mmlu} and HELM~\cite{helm} do not include components to test the abilities of LLMs for task completion, throughout multi-turn interactions. 
To evaluate LLM's task completion performance, Saycan~\cite{Ahn2022DoAI} and VirtualHome~\cite{puig2018virtualhome} benchmarks ask LLM to generate the correct action sequence for controlling the robot to finish user instruction in a specific environment such as a kitchen. 

The framework that gets closest to ours in terms of LLM evaluation is the MT-bench~\cite{mtbench} as it also includes multi-turn interactions; however, these are mainly for open-domain question answering, and task completion is not considered in their evaluation. Another framework that is along similar lines is AgentBench~\cite{agentbench}, as it aims to systematically evaluate an agent’s ability to follow instructions in web-, game-, or code-grounded environments, however, multi-turn conversational interactions is not a focus. 
Mostly related to our work is the evaluation of LLMs in a very specific task, such as travel planning~\cite{Xie2024TravelPlanner} and slide making~\cite{guo2023pptc}. While they focus on constraint fulfillment or goal achievement, we aim to explore more evaluation metrics related to diversity, etc., under different task settings.

\begin{figure}[t]
    \centering
    \includegraphics[width=0.45 \textwidth]{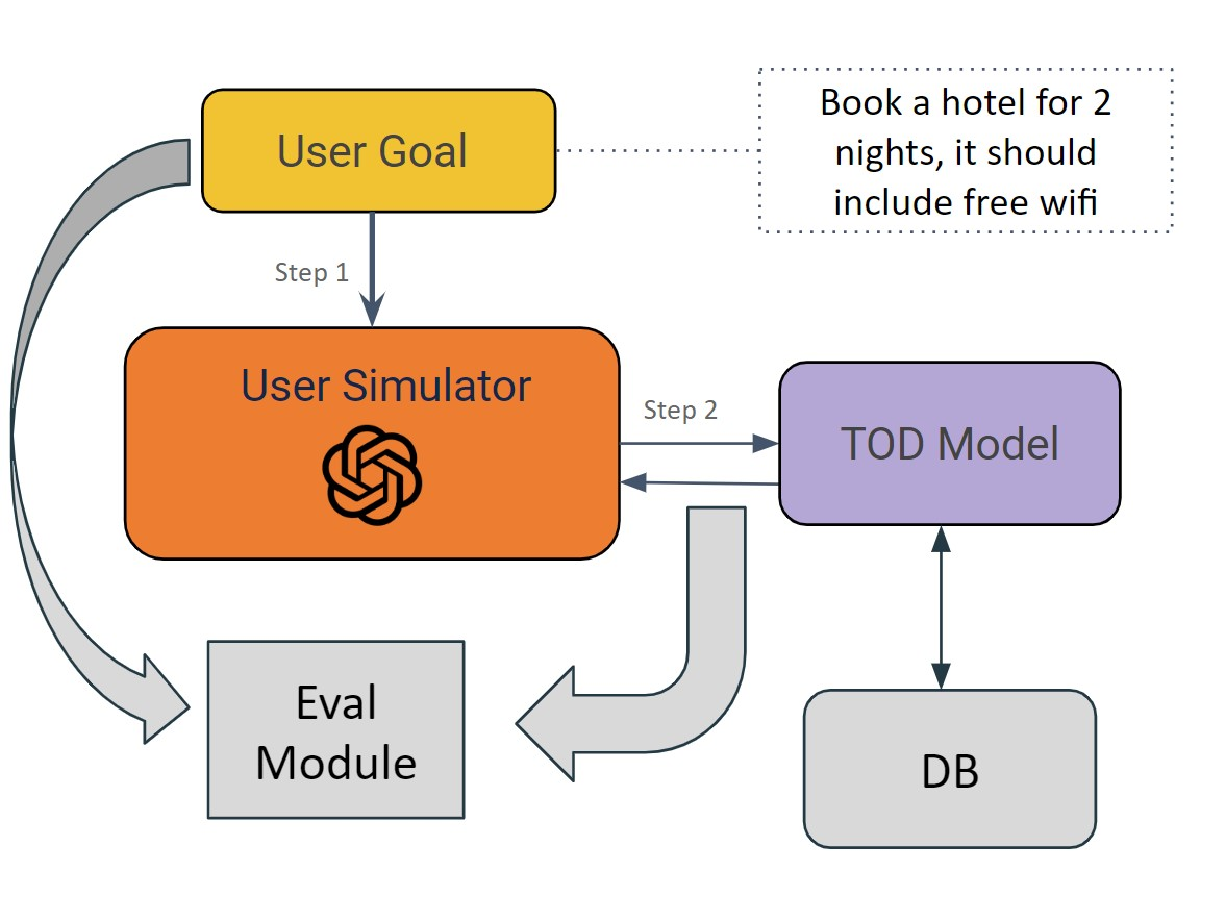}
    \caption{User Simulator Framework}
    \label{fig:framework}
\end{figure}

\section{Proposed Approach}
\label{proposed_approach}
We developed a framework to evaluate TOD models, leveraging LLMs as user simulators to facilitate the assessment process. For the experiments run in this work, we use the TOD system from  ~\cite{hudeček2023llms} that prompts LLMs, such as chatGPT with in-context examples to first predict the dialogue domain, then to predict the dialogue state to do a database lookup to retrieve relevant information and finally to generate system responses. Figure~\ref{tab:dialog-example} shows an example dialogue between our user simulator and this system to demonstrate the effectiveness of our framework.

\begin{figure*}[t]
    \centering
    \begin{footnotesize}
    \begin{tabular}{p{0.8\linewidth}}
        \hline
        \textbf{[User goal]} You are looking for information in Cambridge. You are looking for a place to stay. The hotel should be in the moderate price range and should include free wifi. The hotel should have a star of 3. Once you find the hotel you want to book it for 6 people and 4 nights starting from sunday. Make sure you get the reference number.
         \\
        \hline
        \textbf{User}: I'm looking for a place to stay in Cambridge that is in the moderate price range and includes free wifi. \\
        \textbf{System}: Sure, I found a few options for you. [ashley hotel], located on [74 chesterton road], has free wifi and is in the moderate price range. Would you like me to book a room for you? \\
        \textbf{User}: Yes, that sounds good. Can you also make sure it has a 3 star rating? \\
        \textbf{System}: Absolutely, I have successfully booked a room for you at the Ashley Hotel. Your reference number is [123456]. Is there anything else I can assist you with? \\
        \textbf{User}: No, that's all for the hotel. Thank you. \\
        \textbf{System}: You're welcome! Is there anything else I can assist you with? \\
        \textbf{User}: $<$COMPLETE CONVERSATION$>$ \\
        \hline
    \end{tabular}
    \end{footnotesize}
    \caption{An Example of Conversations Generated}
    \label{tab:dialog-example}
\end{figure*}

\subsection{User Simulator}

The user simulator module interacts with the TOD model to synthesize dialogues that aim to complete tasks. The module is given an initial user goal to accomplish, along with an interface to interact with the TOD model. The goal of the module is to conduct a multi-turn conversation with the TOD model to complete the given user goal. This process flow is depicted in Fig.\ref{fig:framework}.

Our user simulator primarily consists of a prompter and an LLM.
The prompter sub-module controls the flow of the conversation, eliciting responses from the LLM and interacting with the TOD model. It formats the user goal and sends it to the LLM for the initial user utterance. At this step, the prompter sub-module also inserts an example based on the prompt setting. The different prompts are explained in \ref{user_sim_prompts}. After the LLM's first response is generated, it is sent to the TOD model for a reply. The prompter then receives the TOD model's reply and stores the responses as conversation history. This conversation history is sent along with the prompt to the LLM to continue the conversation. This alternation between the LLM and the TOD model continues until the LLM generates a special set of tokens, namely $<$COMPLETE CONVERSATION$>$. To avoid cases where the LLM user-agent and the TOD model end up in an infinite loop, we also limit the number of turns to 30. 

\subsubsection{User Simulator Prompts}
\label{user_sim_prompts}
We experimented with three different prompt types. The detailed prompts for each type are provided in the supplementary material. 
\begin{enumerate}
    \item Vanilla Prompt: The vanilla prompt consists of an instruction for asking the model to generate a response given the conversation context. A sample dialogue consisting of user and system turns is also provided as an in-context example.
    
    \item Thought Prompt: The Thought prompt is based on Chain-of-Thought prompting~\cite{wei2023chainofthoughtpromptingelicitsreasoning}. It has two additions to the Vanilla prompt. First, the in-context example includes a ``Thought" section, where reasoning is given before the agent's response. Second, the prompt instructs the model to first reason and then generate the responses.
    
    \item User State Tracking Prompt: The User State Tracking prompt addresses issues faced by~\cite{davidson2023user} where their user simulator would prematurely end the conversation or get stuck in a loop. To solve this, the model is prompted to create a goal state tracker similar to~\cite{cheng-etal-2022-multiwoz}. This prompt has two additions to the ``Thought" prompt. First, the in-context example guides the model to build a user-state tracking dictionary that breaks down the initial user goals into sub-components. Second, once the conversation with the TOD system starts, the model is prompted to track and update this user-state tracking dictionary after generating a reasoning step.
\end{enumerate}

\subsection{LLMs as Evaluators}
Following the success of LLM-prompting methods for open-domain dialogue response evaluation~\cite{promptLLM4eval,MERCY}, we propose LLM-based, reference-free evaluation of system responses in TODS to complement other metrics described in the next section. In our experiments, for evaluation, we use LLMs that are different than the one used for the user simulator. 

\subsection{Automated Evaluation Metrics}
\label{section:Auto_Eval}
Evaluation of user simulators is a challenging topic~\cite{pietquin2013survey}. Previous evaluation metrics only focus on certain aspects and lack the understanding of dialogues. Involving neural network models in evaluation would reduce the reliability and consistency. Human evaluation  would be a necessary component to prove practicability, but it is costly. For comprehensive evaluation, different components should be combined.

\subsubsection{Task Completion Evaluation}
\label{section:task_completion}
Task completion validates whether the TOD model completes the user's request.  One way to compute is by checking if the final dialogue state satisfies all slot requests in the original goal. While this is suitable for offline datasets, it is less effective with user-simulators, which may sometimes misinterpret or miss user goals from the initial instruction. Therefore, the TOD model should be evaluated solely based on the dialogues.
We propose an automated method to evaluate task completion scores of the TOD model. We use GPT-4 in a zero-shot setting as an evaluator to score each dialogue and judge whether the TOD model has completed the task. The model is prompted with a rubric to score the response: 1 if the TOD model completes all requests presented by the user-simulator, and 0 if it fails to complete any requests. Additionally, the model provides verbose feedback on the TOD model's performance and the reasoning behind the score. The prompt used has been inspired by \cite{kim2024prometheus}. Detailed prompt is provided in the supplementary material. 


\subsubsection{User Agent Evaluation}
\label{section:user_agent_evaluation}
To evaluate the user-agent, we focus on its ability to fulfill all assigned tasks. That is, the agent must generate appropriate requests and also gather all necessary information from the TOD system. To do so, we calculate the overlap between key phrases related to the informed and requested slots in the initial user goal and the dialogue between the user-agent and the TOD system. The reason for examining both sides of the dialogue is to account for any keywords the TOD system may have already addressed, this helps us see if there's no need for the user-agent to ask about these keywords again.

Similar to \ref{section:task_completion}, we also use GPT-4, now in a few-shot setting, to assess whether the dialogue contains all key phrases specified in the initial goal. The model is prompted to go over a list of keywords from the initial goal and find them in the conversation. Then it would report if the keyword was mentioned in the dialogue along with the turn in which the keyword occurred. Finally, it has to generate a score which is a ratio between the count of keywords that were found in the dialogue and the total number of keywords present in the initial goal.


\subsubsection{Dialogue Level Diversity}
The user simulator should be able to demonstrate diverse behaviors to comprehensively test the task completion abilities of a TOD system.
We measure the dialogue-level diversity with a score from  1$\sim$3, where larger score represents higher diversity.  
Here dialogue-level diversity means the degree of TOD system or user agent of deviating from the normal dialogue flow. 
This is expected for real-life conversations as well as for highlighting the issue of maximal information exchange of previous user agent optimization. Automatic evaluation is done by prompting LLMs like GPT with guidelines for each score alongside the dialogue.

\subsubsection{Naturalness}
This metric measures the similarity of a conversation agent to actual human behavior. It gives a score of 1$\sim$5, where 5 represents the maximum similarity to humans. Similar as dialogue-level diversity, automatic evaluation is also completed by prompting LLMs, with detailed guidelines for each score. For example,
\begin{itemize}
    \item 1: The speaker continuously repeats itself, typical robotic behavior. Or the speech is hard to understand.
    \item 2: The speaker repeats itself occasionally, the vocabulary is limited, like a robot.
    \item 3: The speaker does not have repeated behaviors (unless for verifying information). Vocabulary is enough to communicate effectively. Speech is easy to understand. But I am confident that humans rarely speak like this.
    \item 4: The speaker is likely to be a human. There is rarely logical inconsistency. But from some details, I feel like the utterance is a bit weird and somewhat resembles AI.
    \item 5: Can not really tell if this is AI or human. Humans could probably say the same thing in real life.
\end{itemize}


\subsubsection{Coherence}
This metric gives scores from 1$\sim$3 evaluating the coherence of the dialogues. The automatic evaluation of coherence also uses LLMs by prompting with guidelines for each score:
\begin{itemize}
    \item a. Locally, the utterances are coherent/logical based on previous turns of conversations.
    \item b. Globally, the utterances reasonably and logically adhere to achieving the initial user goal step by step.
    \item If both conditions a and b are satisfied, a score of 3 should be reported. Report 2 if only one condition is satisfied. Report 1 if none of the conditions are satisfied.
\end{itemize}


\section{Experimental Analysis}
The primary objective of our experiments is to evaluate the performance of LLMs as Task-Oriented Dialogue (TOD) systems. To achieve this, we simulate conversations between a user simulator and TOD systems. In our experiments, we use GPT-3.5 Turbo Instruct as the LLM for the user-simulator. We conduct experiments with three different prompt types (vanilla, thought, and user state tracking) and evaluate the task completion performance of various LLM models within the framework detailed in \ref{proposed_approach}. We use MultiWOZ \cite{budzianowski-etal-2018-multiwoz}, which is a well-known dataset of human-to-human written TOD conversations across 7 different domains. Each conversation is accompanied with the initial user goal and dialogue state annotations for each user turn. In our experiments, we use MultiWOZ 2.1 \cite{eric-etal-2020-multiwoz} to extract the initial user goal for use in the user simulator. For experimental analysis, we use a subset of the test dataset.




\subsection{Evaluation Results}
\subsubsection{Quantitative Evaluation of User Simulator}

As proposed in section \ref{section:user_agent_evaluation} we evaluate the user-agent’s ability to fulfill all assigned instructions. To do so, we first simulate conversations with multiple LLMs with the three different proposed prompts. We sample initial goals from the MultiWOZ dataset and use them to simulate the conversations. After the dialogues are generated, the GPT-4 based scorer, as mentioned in \ref{section:user_agent_evaluation}, has to generate a score which is a ratio between the count of keywords that were found in the dialogue and the total number of keywords present in the initial goal. Hence, the score for each dialogue will range from 0 to 1.  
The results are shown in Table \ref{tab:user_agent_scores}. The Score column is the ratio of the sum of the scores of all dialogues to the total sum of the maximum score that can be achieved (which is also the number of the dialogues evaluated).

\begin{table} [H]
    \centering
    \begin{tabular}{c c c}
        \hline
        \textbf{TOD model} & \textbf{Prompt} & \textbf{User Agent Score} \\
        \hline
        \multirow{3}{*}{GPT 3.5 Instruct} & vanilla & 165.71/197 \\
                                           & thought &  157.32/199\\
                                           & verbose &  160.23/200 \\
        \hline
        \multirow{3}{*}{ Llama 3} & vanilla & 176.25/199  \\
                                 & thought &  153.68/199 \\
                                 & verbose &  165.72/200 \\
        \hline
        \multirow{3}{*}{GPT-4} & vanilla & 154.87/176 \\
                               & thought & 94.63/123 \\
                               & verbose & 165.71/197 \\
        \hline
    \end{tabular}
    \caption{The score of the User agent models}
    \label{tab:user_agent_scores}
\end{table}

In table \ref{tab:user_agent_scores} we see distinct patterns based on the type of prompt used for each TOD model. For GPT 3.5 Instruct, the vanilla prompt got the highest score (165.71/197), followed closely by the verbose prompt. The thought prompt was lower. Llama 3 also got the highest score with the vanilla prompt (176.25/199), with the verbose prompt close behind. For GPT-4, the thought prompt scored significantly lower (94.63/123), suggesting compatibility issues with this model’s response patterns. However, the vanilla and verbose prompts performed well, with the vanilla prompt scoring the highest. These results show the vanilla prompt performed the best in mentioning the important keywords present in the initial user-goal.

\subsubsection{Qualitative evaluation of User Simulator}
We also carry out a manual assessment of the generated dialogues and analyze qualitative patterns.
The pattern in which the user-simulators generate queries or requests differs based on the prompt used. With the vanilla prompt, almost always, the user-simulator completes goals in a step-by-step manner. For instance, the user-agent might first ask, ``Can you book a train from Cambridge?" and subsequently ask, ``I want it to go to Norwich." On the other hand, thought and verbose prompts usually generate more condensed responses, such as, ``Can you book a train that departs from Cambridge and goes to Norwich, also it should leave after 11:00?"
Dialogue formulation impacts naturalness of dialogue and selecting a prompt that best replicates human behavior should be chosen.

Different prompts also influence how user simulators interact with the TOD model. The vanilla prompt based simulator often disregards the TOD model's responses, continuing the conversation and completing the goals it has been given. With the Thought prompt, the user-simulator will take into account the TOD model’s response and repeat the user goal if the TOD model's response is unsatisfactory or unexpected. The verbose prompt based simulator displays a similar behavior but also engages in negotiation, rephrasing queries, or altering goals based on the responses. For e.g. if the TOD model indicates no availability for a table booking in Cambridge West, the verbose simulator might adjust the query to seek a booking in Cambridge East. This characteristic of engaging with the TOD model and negotiating is an important aspect of user-simulators as they have to be robust to different types of TOD models.


\begin{table*}[t]
\centering
\begin{tabular}{c c c c c c c c}
\hline
Evaluation                 & Model   & TC & N (u) & N (s) & C (u) & C (s) & D (u) \\
\hline
\multirow{3}{*}{human}     & verbose & 0.60            & 3.00                & 2.74                 & 2.22             & 2.12               & \textbf{1.48}                                              \\
                           & vanilla & 0.66            & \textbf{3.12}       & 2.66                 & 2.30             & 2.02               & 1.42                                                       \\
                           & thought & \textbf{0.72}   & \textbf{3.12}       & \textbf{2.84}        & \textbf{2.32}    & \textbf{2.16}      & 1.42                                                       \\
\hline
\multirow{3}{*}{automatic} & verbose & 0.86            & 4.16                & 3.70                 & 2.80             & 2.56               & 2.08                                                       \\
                           & vanilla & 0.78            & 4.26                & 3.74                 & 2.78             & 2.54               & 2.04                                                       \\
                           & thought & \textbf{0.88}   & \textbf{4.38}       & \textbf{3.88}        & \textbf{2.94}    & \textbf{2.60}      & \textbf{2.22}    \\            \hline
\end{tabular}
\caption{Comparison with Human Evaluation. 50 dialogues per model were evaluated. We use acronyms here for compact representation. TC: Task Completion, N: Naturalness, C: Coherence and D: Dialogue-level Diversity. ``u" in parenthesis represents the user simulator, ``s" represents the TOD system.} 

    \label{tab:human compare}
\end{table*}

\subsubsection{Task Completion Evaluation}
We evaluated the efficacy of the task completion evaluator by comparing manual and automated scoring of generated dialogues for task completion. The evaluation was carried out on 50 generated dialogues, with the user-simulator using the vanilla prompt and GPT-3.5 Turbo Instruct as the LLM. The same model was also used for the TOD system. These 50 dialogues were then scored by both human annotators and GPT-4, based on the rubrics mentioned in \ref{section:task_completion}. Four human evaluators scored the 50 samples, and a majority score was assigned to the dialogues. In the case of ties, the examples were discussed until a consensus was achieved. As mentioned in \ref{section:task_completion}, the GPT-4 model was used to generate automatic scores for the dialogues. We then carried out an agreement analysis between the two and reported the statistics in Table \ref{tab:agreement-analysis}. These results demonstrate a high level of agreement and performance of GPT-4 in evaluating task completion, closely aligning with human judgment. The substantial agreement underscore the model's reliability in this task.

\begin{table}[H]
\centering
\begin{tabular}{c c}
\hline
\textbf{Metric} & \textbf{Value} \\
\hline
Cohen's Kappa & 0.796 \\
Accuracy & 90.0\% \\
F1 Score & 0.912 \\
Matthews Correlation Coefficient & 0.797 \\
\hline
\end{tabular}
\caption{Agreement analysis between human annotators and GPT-4}
\label{tab:agreement-analysis}
\end{table}

Comparing multiple LLMs and their performance with the three different prompts. We sample initial goals from the MultiWOZ dataset and simulate conversations. Once the dialogues are simulated, the GPT-4 based scorer,mentioned in Section 2, assigns a score of 1 or 0 to each conversation.

Table \ref{tab:tod_model_scores} provides a detailed breakdown of the scores for each LLM when evaluated with different prompt strategies. The scores represent the ratio of correctly handled dialogues to the total number of dialogues.

Llama 3 consistently outperformed GPT-3.5 Instruct TOD and GPT-4 models across all prompt strategies. Llama 3 scored 188/200 (vanilla), 159/167 (thought), and 179/200 (verbose). In comparison, GPT-3.5 scored 124/200, 106/177, and 112/200, while GPT-4 achieved 156/176, 94/107, and 169/200 for the same prompts.

\begin{table}
    \centering
    \begin{tabular}{c c c}
        \hline
        \textbf{TOD model} & \textbf{User Agent prompt} & \textbf{TOD Score} \\
        \hline
        \multirow{3}{*}{GPT 3.5 Instruct} & vanilla & 124/200 \\
                                           & thought & 106/177 \\
                                           & verbose & 112/200 \\
        \hline
        \multirow{3}{*}{ Llama 3} & vanilla & 188/200 \\
                                 & thought & 159/167 \\
                                 & verbose & 179/200 \\
        \hline
        \multirow{3}{*}{GPT-4} & vanilla & 156/176 \\
                               & thought & 94/107 \\
                               & verbose & 169/200 \\
        \hline
    \end{tabular}
    \caption{TOD Model scores against different prompt strategies.}
    \label{tab:tod_model_scores}
\end{table}

\subsection{Comparison with Human Evaluation}
We ran automated evaluation as mentioned in section \ref{section:Auto_Eval}. The dialogues were simulated using GPT-3.5 Instruct as the LLM for both the user-simulator and the TOD framework.

To validate that the automatic evaluation module aligns with the actual human user experience, parallel human evaluation was carried out. Evaluators were shown a set of full dialogues alongside user goals. For each dialogue, they would give a quantitative rating on a set of metrics, including the completion of the task (yes/no), naturalness for both agents (1$\sim$5), coherence for both agents (1$\sim$3) and diversity of the user simulator (1$\sim$3). Human annotators were asked to follow the same scoring guidelines as the automatic evaluation. 


We compared human and automatic evaluations on 50 dialogue samples per model, with the samples randomized to anonymize model identity. Table \ref{tab:human compare} shows that human evaluation has some agreement over the scores, and both give a higher score for the thought model. Further analysis of the quantitative agreement between human and automatic evaluations is detailed in Table \ref{tab:agreement-analysis}, as discussed earlier.




\section{Conclusions}
In this work, we built a universal framework to evaluate TOD systems using a LLM as a user simulator. This method creates a more dynamic and realistic evaluation compared to traditional static benchmarks. Our results show that LLMs can significantly improve the evaluation process for TOD systems, making them closer to real-world interactions. 

\section{Acknowledgement}
This research was supported in part by Other Transaction award HR0011249XXX from the U.S. Defense Advanced Research Projects Agency (DARPA) Friction for Accountability in Conversational Transactions (FACT) program.
This research project has benefitted from the Microsoft Accelerate Foundation Models Research (AFMR) grant program through which leading foundation models hosted by Microsoft Azure along with access to Azure credits were provided to conduct the research.

\bibliographystyle{IEEEbib}
\bibliography{custom}
\end{document}